\documentclass[letterpaper]{article} 
\usepackage{aaai2026_arxiv}  
\usepackage{times}  
\usepackage{helvet}  
\usepackage{courier}  
\usepackage[hyphens]{url}  
\usepackage{graphicx} 
\usepackage{natbib}  
\usepackage{caption} 
\usepackage{pdfpages}
\usepackage{algorithm}
\usepackage{algorithmic}
\usepackage{subfig}

\usepackage{newfloat}
\usepackage{listings}

\usepackage[pagebackref,breaklinks,colorlinks]{hyperref}
\usepackage[table]{xcolor}

\usepackage{graphicx}
\usepackage{amsmath}
\usepackage{amssymb}
\usepackage{booktabs,siunitx}
\usepackage{bbding}
\usepackage{pifont}%
\usepackage{multirow}

\usepackage[capitalize]{cleveref}
\crefname{section}{Sec.}{Secs.}
\Crefname{section}{Section}{Sections}
\Crefname{table}{Table}{Tables}
\crefname{table}{Tab.}{Tabs.}

\usepackage[table]{xcolor}
\usepackage{makecell}

\usepackage{xcolor}
\usepackage{colortbl}
\definecolor{baselinecolor}{gray}{.9}

\definecolor{color1}{RGB}{113,150,108}
\definecolor{color2}{RGB}{146,117,145}
\definecolor{color3}{RGB}{235,159,131}
\definecolor{color4}{RGB}{108,141,165}
\definecolor{color5}{RGB}{229,154,128}
\definecolor{color6}{RGB}{104,137,159}
\definecolor{color7}{RGB}{250,227,219}

\DeclareCaptionStyle{ruled}{labelfont=normalfont,labelsep=colon,strut=off} 
\floatstyle{ruled}
\newfloat{listing}{tb}{lst}{}
\floatname{listing}{Listing}
\title{A Study of Finetuning Video Transformers for Multi-view Geometry Tasks}
\author {
    Huimin Wu\textsuperscript{\rm 1},
    Kwang-Ting Cheng\textsuperscript{\rm 1},
    Stephen Lin\textsuperscript{\rm 2},
    Zhirong Wu\textsuperscript{\rm 2}
}
\affiliations {
    \textsuperscript{\rm 1}The Hong Kong University of Science and Technology\\
    \textsuperscript{\rm 2}Microsoft Research Asia\\
    hwubl@connect.ust.hk, 
    timcheng@ust.hk, 
    stevelin@microsoft.com,
    wuzhiron@microsoft.com \\
    \textbf{Project website}: \href{https://geovit-aaai26.github.io}{geovit-aaai26.github.io}
}

\usepackage{bibentry}

\begin{document}

\maketitle
\begin{abstract}


This paper presents an investigation of vision transformer learning for multi-view geometry tasks, such as optical flow estimation, by fine-tuning video foundation models.
Unlike previous methods that involve custom architectural designs and task-specific pretraining, 
our research finds that general-purpose models pretrained on videos can be readily transferred to multi-view problems with minimal adaptation.
The core insight is that general-purpose attention between patches learns temporal and spatial information for geometric reasoning.
We demonstrate that appending a linear decoder to the Transformer backbone produces satisfactory results, and iterative refinement can further elevate performance to state-of-the-art levels.
This conceptually simple approach achieves top cross-dataset generalization results for optical flow estimation with end-point error (EPE) of 0.69, 1.78, and 3.15 on the Sintel clean, Sintel final, and KITTI datasets, respectively. 
Our method additionally establishes a new record on the online test benchmark with EPE values of 0.79, 1.88, and F1 value of 3.79.
Applications to 3D depth estimation and stereo matching also show strong performance,
illustrating the versatility of video-pretrained models in addressing geometric vision tasks. 

\end{abstract}
\section{Introduction}

There exists considerable interest in building general-purpose vision foundation models~\cite{he2022masked,alayrac2022flamingo,wang2023internimage} capable of addressing a range of tasks from low-level reasoning to high-level understanding. 
However, their performance on the downstream tasks remains inferior to methods with elaborate architectural designs and task-specific pretraining. Further research is needed to fully realize the potential of vision foundation models for various applications.

For multi-view geometry tasks such as optical flow estimation, complicated pipelines are generally employed to enhance performance. For example, mainstream optical flow estimators utilize a convolutional stem to extract per-frame features, a transformer encoder to extract a cost volume as an explicit motion representation, and a recurrent neural network to conduct iterative refinement~\cite{teed2020raft,jiang2021learning,huang2022flowformer,shi2023flowformer++}. The state-of-the-art method Flowformer++~\cite{shi2023flowformer++} additionally uses a domain-specific pretraining scheme.
Towards simplifying the pipeline, CroCo~\cite{weinzaepfel2022croco} and its extension~\cite{weinzaepfel2023croco} use  transformers as well as an adaptation module for dense prediction~\cite{ranftl2021vision}. However, this does not provide a general solution because of its reliance on a specialized pretraining task, an in-house dataset obtained via meticulously controlled data collection, and a much larger downstream dataset than used in common practice.

In this work, we investigate a \textit{simpler and more general} solution for multi-view geometry tasks. Our approach is guided by the intuition that cross-frame self-attention in video-pretrained vision transformers can capture matching information essential for these multi-view problems. Based on this idea, we adapt general-purpose ViTs pre-trained on videos~\cite{feichtenhofer2022masked,wang2022internvideo} to multi-view geometry tasks with minimal adjustment. After modifying pre-trained 3D ViTs to accommodate two-frame tasks, we found that appending a linear head is sufficient to obtain high performance, and iteratively refining the results further enhances prediction quality.

Our model, called \textbf{GeoViT}, was evaluated on the tasks of optical flow estimation, stereo matching and 3D depth estimation.
For optical flow estimation, it sets a new standard on the challenging Sintel (clean and final splits) and KITTI datasets, 
reducing EPE by 21\%, 9.6\%, and F1 by 11.2\%, respectively.
For 3D depth estimation, GeoViT achieves state-of-art results on the RGBD\_SLAM, SUN3D and Scenes11 datasets across multiple evaluation metrics.
For stereo matching, our model outperforms competing methods on the ETH3D dataset for two of the three metrics.

This work makes an important contribution to the research community by demonstrating that the robust feature representations learned during video pre-training capture temporal and spatial information useful across various geometric tasks. It presents a simple yet highly effective way to harness this knowledge without custom pretraining strategies or special data collection. We believe that the connection between video understanding and geometric computer vision has been underleveraged, and we hope that our work will inspire further geometric-based research that capitalizes on the rich representations learned by video-based models. 

\section{Related Works}

\noindent
\textbf{Foundation models.}
The rise of self-supervised learning has enabled vision models~\cite{wu2018unsupervised,chen2020simple,he2020momentum,he2022masked} to capture rich semantics from vast raw data without human labels. These pretrained models, when finetuned to individual downstream tasks~\cite{li2022exploring, lin2023detr,liu2023simpleclick,xu2022vitpose}, have consistently shown improved generalization and have sparked interest in building general-purpose foundation models to solve all computer vision applications within a single model.

Large-scale language models (e.g., GPT~\cite{brown2020language}) and multi-modal models (e.g., CLIP~\cite{radford2021learning}) demonstrate strong performance using language as supervision for self-supervised learning.
More importantly, language as a natural interface between applications allows the foundation model to perform tasks in a zero-shot or few-shot manner.
Recent vision foundation models~\cite{wang2022internvideo,wang2023internimage,liu2024visual,bai2024sequential} follow this trend by integrating language and formulating heterogeneous supervisions. 
Despite the flexibility of the language interface, vision foundation models often need to be finetuned in order to reach optimal performance.

The promise of foundation models lies in their generality and scalability. Nevertheless, they still fall short in performance compared with domain-specific methods. Unleashing the full potential of foundation models for end tasks has become a problem of great interest.

\noindent
\textbf{Optical flow estimation.} 
Optical flow estimation predicts per-pixel displacement from the source to the target image.
Its widespread use across various vision-related tasks makes it a fundamental element of computer vision. 
For example, flow estimates can offer additional supervisory cues into tasks ranging from action recognition~\cite{simonyan2014two,sun2018optical} and tracking~\cite{schulter2017deep,zheng2021learning} to video segmentation~\cite{cheng2017segflow,xiao2018monet,xu2018dynamic} and self-supervised learning~\cite{huang2023mgmae}.

Methods for predicting optical flow commonly build a cost volume~\cite{dosovitskiy2015flownet} to characterize the dense correlation between pixels. 
Feature enhancement~\cite{sun2022skflow,sui2022craft,huang2022flowformer,shi2023flowformer++} for the cost volume further improves the prediction accuracy. Iterative refinement~\cite{teed2020raft,huang2022flowformer} is another crucial technique for addressing flows with large displacement, where the prediction at the current timestep is fed back to the model as inputs for refinement. 

A line of work~\cite{xu2022gmflow, xu2023unifying,weinzaepfel2022croco, weinzaepfel2023croco} attempts to simplify the pipeline using generic transformers for optical flow estimation. 
As transformers lack inductive biases, the model needs to be pretrained with custom pretext tasks on particular types of datasets. However, these methods have yet to reach the performance levels of heavily engineered approaches.

\noindent
\textbf{Stereo matching.}
Predicting disparity for stereo matching involves estimating the horizontal pixel displacement between the left and right stereo images. 
Given the similarities with flow prediction, 
recent advancements~\cite{lipson2021raft} have adapted the flow network RAFT~\cite{teed2020raft} for stereo applications, 
where a 3D correlation volume is constructed instead of a 4D cost volume due to epipolar constraints. 
Furthermore, improvements have been made to enhance the correlation volume's robustness to non-ideal rectification~\cite{li2022practical} and its representative ability by introducing non-local information~\cite{xu2023iterative}.
Additionally, efforts have been made to tame transformers to improve stereo matching~\cite{guo2022context,li2021revisiting}.

\noindent
\textbf{Depth estimation.} 
Our work is applied to unrectified multi-view depth estimation, which predicts depth from images captured without rectifying them to a common plane.
This setting allows for more natural and diverse camera configurations.
A commonly used method is plane-sweep stereo~\cite{collins1996space, im2019dpsnet}, 
which constructs a 3D cost volume where the third dimension represents potential depth hypotheses. 
Subsequent improvements have enhanced the cost volume design by improving the multi-scale representation of depth information~\cite{yang2020cost, gu2020cascade} and by using transformers to better capture global context~\cite{ding2022transmvsnet,xu2023unifying}.


\section{Approach}

\begin{figure*}[t]
\begin{center}
   \includegraphics[width=0.9\linewidth]{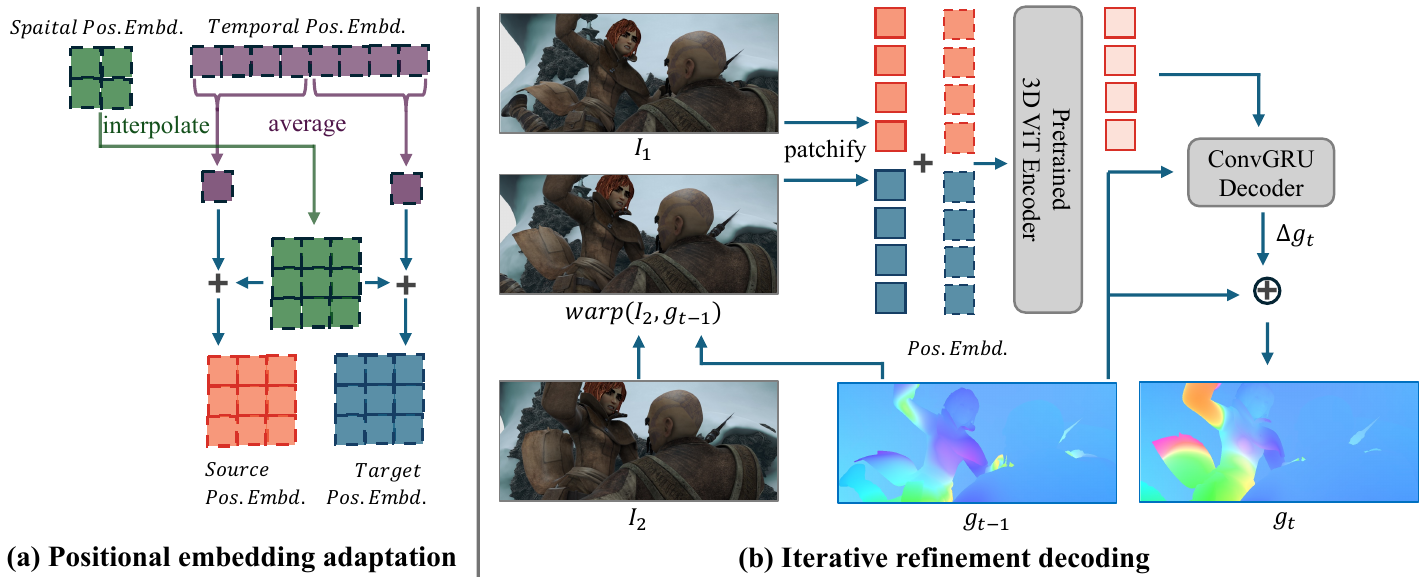}
\end{center}
   \caption{Overview of GeoViT. 
   Part (a) presents adaptation of positional embeddings in pretrained 3D ViTs. The pretrained spatial Pos. Embd. (in \colorbox{color1}{green}) are interpolated to match the desired input size. The pretrained temporal Pos. Embd. (in \colorbox{color2}{purple}), which accounts for 8 frames, is split into two halves.  The average of the first half and second half corresponds to the temporal embedding of the source image (in \colorbox{color3}{orange}) and target image (in \colorbox{color4}{blue}), respectively. Part (b) exhibits our iterative refinement decoding pipeline, using optical flow for illustration. The input target image is dynamically warped based on the last-step prediction $g_{t-1}$ so that the input pair corresponds to the ground truth residual for this step. Then the source and (warped) target images are patchified, added with adapted positional embeddings, and fed to the pretrained 3D ViT for feature extraction. The decoder accepts source image features (in \colorbox{color7}{light orange}) and the last-step prediction ($g_{t-1}$) and produces its correction $\Delta g_t$. Adding $g_{t-1}$ and its correction gives the current-step prediction $g_{t}$.
   }
\label{fig:overview}
\end{figure*}

We explore the use of large-scale video foundation models for geometric vision tasks, including optical flow, stereo correspondence, and depth estimation. Our intuition is that the cross-frame attention mechanisms learned by transformer-based architectures encode not only high-level semantic relationships but also rich correspondence cues that can be leveraged to recover fine-grained, low-level matches.


To this end, we propose GeoViT, a unified framework that takes an image pair as input and predicts the geometric quantity relating the two views
This formulation enables optical flow, stereo matching, and depth estimation to be addressed within a single architecture.
GeoViT is built around a pre-trained spatiotemporal Vision Transformer (ViT) encoder, which is demonstrated to provide strong geometric representations.
By further integrating within an iterative refinement loop, the framework captures fine-grained geometric relationships while incrementally improving estimation accuracy through feedback and adjustment.

In this section, 
we will first briefly describe video foundation models based on vision transformers.
Following this, 
we will discuss our simple yet effective adaptation scheme designed for downstream geometric tasks, 
emphasizing that the iterative refinement strategy empowers GeoViT to achieve superior results. 

\subsection{Pretrained Video Foundation Models} 

Our work transfers the encoder in video foundation models to multi-view geometry tasks.
Any video foundation model whose encoder follows a transformer architecture can be used.
In our study, the models include MVD~\cite{wang2023masked}, InternVideo~\cite{wang2022internvideo}, UMT~\cite{li2023unmasked}, MAE~\cite{he2022masked}, and MAE\_st~\cite{feichtenhofer2022masked}.
To process the video data, the transformer splits the spatio-temporal data into 3D patches, adds spatial and temporal positional encodings, and then feeds the visual tokens into the self-attention blocks.

Video foundation models are primarily benchmarked on semantic reasoning tasks, 
such as video action recognition~\cite{feichtenhofer2022masked, tong2022videomae} and video action detection~\cite{wang2023videomae}. 
Since the video encoder is trained on multi-frame input while most geometric tasks process a pair of frames, the patchification and position encodings need to be adjusted.  

To adapt pretrained 3D ViTs for two-frame tasks, 
we first interpolate the 2D spatial positional encodings to match the desired input size.
The temporal dimension in the pretrained 3D patch embeddings is usually 8 frames. 
We split the temporal position encodings into two halves and calculate the average of each half along the temporal dimension.
For the patch embedding, we sum the 3D patch embedding along the temporal dimension and convert it basically to a 2D patch embedding.
In this way, we adjust the position encodings and patch embeddings for the pair-wise geometry tasks.
The final token representation is the addition of 2D patch embedding, 2D positional encoding, and temporal encoding.
This approach ensures that the model effectively captures the necessary spatial and temporal information while maintaining compatibility with the pretrained model.



\subsection{Simple Linear Decoding}

A key driver behind our approach is the inherent ability of transformers to model dense correlations,
making it particularly well-suited for per-pixel predictions in downstream geometric applications. 
Moreover, the self-attention mechanism features a full-context receptive field, 
which significantly expands the range of contextual information it can capture.
This design is in line with recent advancements that focus on enhancing feature interactions across both spatial dimensions~\cite{jiang2021learning, sun2022skflow} and motion dimensions~\cite{xu2022gmflow, xu2023unifying, shi2023flowformer++}.

As an initial step, we append a linear layer to each output patch representation, and directly regress the geometrical properties. 
With this minimal modification to the pre-trained encoder, strong performance can already be achieved as shown in Table~\ref{tab:abl_iter}. For optical flow estimation, 
our method attains an End Point Error (EPE) of 2.0 on the Sintel (final) dataset, 
surpassing existing state-of-the-art methods like SAMFlow (2.11).
This demonstrates that the transformer encoder learned from vast video data effectively learns patch-wise correlations useful for geometrical reasoning.

\subsection{Iterative Refinement Decoding}

To further enhance performance,
we propose to incorporate an iterative refinement mechanism inspired by RAFT~\cite{teed2020raft}.
In RAFT, the optical flow prediction is decomposed and serialized to a sequence of residuals where a recurrent neural network (RNN) is used to predict the residual at each timestep. 
The summation of the residuals corresponds to the final prediction.
The input feature at each timestep to the RNN is dynamically constructed to facilitate residual prediction. Specifically, for each source pixel, it queries the cost volume based on the current optical flow estimate.
While it is feasible to migrate this refinement approach to vision transformer backbones, this would face two issues: 1) the cost volume construction is complex in design and overlaps the modeling role of self-attentions, 2) the lookup operation at each timestep is local and suboptimal~\cite{huang2022flowformer, shi2023flowformer++}.

To overcome the aforementioned limitations, 
we propose to remove reliance on the cost volume for a simple yet effective refinement framework.
Formally, given an image pair $I_1, I_2$, the residual geometric property $\Delta g_t$ at each iteration $t$ is predicted as follows:

\begin{equation}
    \Delta g_t = F_{\text{dec}}(F_{\text{enc}}(I_1, \text{warp}(I_2, g_{t-1})), g_{t-1}),
\end{equation}
\begin{equation}
g_t=g_{t-1}+\Delta g_t,    
\end{equation}
where $F_{\text{enc}}$ denotes a spatiotemporal ViT that takes (warped) image pairs as input and returns features corresponding to source images.
The source image features together with the current prediction are fed to the decoder $F_{\text{dec}}$ to predict the residuals.
The decoder is instantiated as a ConvGRU unit following RAFT.
The warping operation takes an image as input and outputs another image according to the geometrical property $g_{t-1}$. The warping operation for optical flow estimation and stereo matching is straightforward. For 3D depth estimation, we first convert the depth representation to pixel displacement using camera parameters, and convert the prediction results back to depth.
The predicted residual is then aggregated with the prediction of the previous step.

The generated sequence of predictions ${g_1, g_2, ..., g_T}$ are regressed to minimize their L1 distance with the ground truth, weighted by exponentially increasing factors.
Formally, the loss function for a pair of images is defined as:
\begin{equation}
    L(I_1,I_2,y) = \sum_{t=1}^{T}\gamma^{T-t}d_{L1}(g_t, y),
\end{equation}
where $y$ represents the ground truth map and $d_{L1}$ denotes L1 distance.
$\gamma$ is the weighting factor that assigns a higher weight to later predictions, 
which is set to 0.9 by default.
$T$ denotes the number of iterations.

The overall iterative refinement uses only a generic pretrained encoder plus a lightweight recurrent neural network. 
It does not rely on customized representations such as a cost volume. 
Looping over the encoder allows the refinement to take greater advantage of large-scale pretrained models.

A concern about the approach is that it adds significant computational complexity. However, we find this not to be a serious issue because our largest model can be trained on just 8 V100 GPUs for at most 2 weeks using techniques such as gradient checkpointing. 
This paper aims to develop a conceptually simple approach. Controlling factors such as data size and model size for fair comparisons is challenging.

\section{Experiments}

In this section, we conduct comprehensive experiments qualitatively and quantitatively for optical flow estimation, stereo matching and 3D depth estimation. Ablation studies are  performed on the optical flow estimation task. Most of the results are submitted to evaluation servers to obtain official results. 
In the following, we describe each task. 

\begin{table}[t]
\centering
\setlength\tabcolsep{1pt}
{
\scriptsize{
\begin{tabular}[t]{cccccc}
\hline
\multirow{2}{*}{ \shortstack{Training\\ Data} } & \multirow{2}{*}{ Method } & \multicolumn{2}{c}{ Sintel (train) } & \multicolumn{2}{c}{ KITTI-15 (train) } \\
\cmidrule(lr){3-4}\cmidrule(lr){5-6}
& & Clean & Final & F1-epe & F1-all \\
\midrule 
\multirow{3}{*}{ A } & Perceiver IO~\cite{jaegle2021perceiver} & 1.81 & 2.42 & 4.98 & -\\
& PWC-Net~\cite{sun2018pwc} & 2.17 & 2.91 & 5.76 & -\\
& RAFT~\cite{teed2020raft} & 1.95 & 2.57 & 4.23 & -\\
\midrule 
\multirow{14}{*}{$\mathrm{C}+\mathrm{T}$} & HD3~\cite{yin2019hierarchical} & 3.84 & 8.77 & 13.17 & 24.0 \\
& LiteFlowNet~\cite{hui2018liteflownet} & 2.48 & 4.04 & 10.39 & 28.5 \\
& PWC-Net~\cite{sun2018pwc} & 2.55 & 3.93 & 10.35 & 33.7 \\
& LiteFlowNet2~\cite{hui2020lightweight} & 2.24 & 3.78 & 8.97 & 25.9 \\
& S-Flow~\cite{zhang2021separable} & 1.30 & 2.59 & 4.60 & 15.9 \\
& RAFT~\cite{teed2020raft} & 1.43 & 2.71 & 5.04 & 17.4\\
& FM-RAFT~\cite{jiang2021learning_2} & 1.29 & 2.95 & 6.80 & 19.3 \\
& GMA~\cite{jiang2021learning} & 1.30 & 2.74 & 4.69 & 17.1\\
& GMFlow~\cite{xu2022gmflow} & 1.08 & 2.48 & 7.77 & 23.40 \\
& GMFlowNet~\cite{zhao2022global} & 1.14 & 2.71 & 4.24 & 15.4 \\
& CRAFT~\cite{sui2022craft} & 1.27 & 2.79 & 4.88 & 17.5 \\
& SKFlow~\cite{sun2022skflow} & 1.22 & 2.46 & 4.47 & 15.5 \\
& FlowFormer~\cite{huang2022flowformer} & ${0.94}$ & 2.33 & ${4.09}^{\dagger}$ & {14.72}$^{\dagger}$ \\
& FlowFormer++~\cite{shi2023flowformer++} & 0.90 & {2.30} & 3.93$^{\dagger}$ & 14.13$^{\dagger}$ \\
& SAMFlow~\cite{zhou2024samflow} & 0.87 & 2.11 & ${3.44}$ & 12.28 \\
&DPFlow~\cite{morimitsu2025dpflow} &1.02&2.26&3.37& \textbf{11.1} \\
& GeoViT & \textbf{0.69}$^{\dagger}$ & \textbf{1.78}$^{\dagger}$ & \textbf{3.15}$^{\dagger}$ & 11.45$^{\dagger}$ \\

\hline

\end{tabular}
}
}
\caption{ 
Experiments on Sintel~\cite{butler2012naturalistic} and KITTI~\cite{geiger2013vision} datasets.
`A' denotes training on the autoflow dataset.
`C + T' denotes training sequentially on the FlyingChairs and FlyingThings datasets.  
$^{\dagger}$ represents evaluation with tiling technique~\cite{jaegle2021perceiver}.
}
\label{tab:optical_flow}
\end{table}

\begin{table*}[t]
\begin{minipage}{1\linewidth}
\centering
\subfloat[
Pretraining schemes. 
Spatiotemporal masked autoencoder~\cite{feichtenhofer2022masked} as flow estimation initialization outperforms its semantics-rich extensions~\cite{wang2023masked, wang2022internvideo, li2023unmasked} and motion-lacking predecessor~\cite{he2022masked}.
\label{tab:abl_pretraining}
]{
\begin{minipage}{0.5\linewidth}
{
\begin{center}
\scriptsize{
\begin{tabular} {ccccc}
\multirow{2}{*}{ Initialization } & \multicolumn{2}{c}{ Sintel (train) } & \multicolumn{2}{c}{ KITTI-15 (train) }\\
& Clean & Final & F1-epe & F1-all  \\
\midrule 
Rand. Init. & 13.49 & 13.49 &  36.74 & 85.35  \\
MVD~\cite{wang2023masked} & 0.84 & 1.91 & 3.61 & 14.28 \\
InternVideo~\cite{wang2022internvideo} & 0.70 & 1.87 & 3.22 & 12.27 \\
UMT~\cite{li2023unmasked} & 0.78 & 1.89 & 3.73 & 14.13\\
MAE~\cite{he2022masked} & 0.78& 1.99 & 4.18 & 14.85\\
MAE\_st~\cite{feichtenhofer2022masked} & \textbf{0.69} & \textbf{1.78} & \textbf{3.15} & \textbf{11.45} \\

\end{tabular}
}
\end{center}}
\end{minipage}
}
\hspace{3em}
\subfloat[
The effect of training iteration number.
The performance of training with more iterations grows until reaching 6 iterations.
\label{tab:abl_iter}
]{
\centering
\begin{minipage}{0.4\linewidth}{\begin{center}
\scriptsize{
\begin{tabular}{ccccc}
\multirow{2}{*}{\makecell[c]{\# Training \\ iterations}} & \multicolumn{2}{c}{ Sintel (train) } & \multicolumn{2}{c}{ KITTI-15 (train) }\\
& Clean & Final & F1-epe & F1-all  \\
\midrule 
1(linear) & 0.91 & 2.00 & 4.93 & 20.47\\
1& 0.95 & 2.04 & 4.78 & 19.85\\
2& 0.80 & 1.92 & 3.68 & 14.42\\
6&  \textbf{0.69} & \textbf{1.78} & 3.15 & 11.45\\
12& 0.70 & \textbf{1.78} & \textbf{3.13} & \textbf{11.29}\\
\end{tabular}}
\end{center}}
\end{minipage}
}
\\
\subfloat[
Model size. A larger model leads to better results. This table uses reproduced pre-training parameters with fewer pre-training steps (800 epochs instead of the default 1600 epochs).
 \label{tab:abl_model_size}
]{
\begin{minipage}{0.43\linewidth}{
\begin{center}
\scriptsize{
\begin{tabular}{ccccc}
\multirow{2}{*}{ Model } & \multicolumn{2}{c}{ Sintel (train) } & \multicolumn{2}{c}{ KITTI-15 (train) }\\
& Clean & Final & F1-epe & F1-all  \\
\midrule 
ViT-Small& 0.99 & 2.32 & 6.33 & 19.81 \\

ViT-Base& 0.87 & 2.22 & 4.72 & 15.81 \\

ViT-Large& \textbf{0.75} & \textbf{1.82} & \textbf{3.70} & \textbf{13.78} \\
\end{tabular}}
\end{center}}
\end{minipage}
}
\hspace{1em}
\subfloat[
The effect of warping images. 
Warping images for refinement performs better than using cost volumes.
\label{tab:abl_decoder}
]{
\centering
\begin{minipage}{0.52\linewidth}{\begin{center}
\scriptsize{
\begin{tabular}{ccccccc}
\multirow{2}{*}{Method}&\multirow{2}{*}{{Features}} &\multirow{2}{*}{{Refinement}} & \multicolumn{2}{c}{ Sintel (train) } \\
& & & Clean & Final\\
\midrule 
RAFT & CNN & Queried cost volume & 1.43 & 2.71 \\
RAFT (ViT context) & Video ViT & Queried cost volume
 & 0.85 & 1.99\\
GeoViT & Video ViT & Image warping &  {\textbf{0.69}} & {\textbf{1.78}} \\
\end{tabular}}
\end{center}}
\end{minipage}
}
\caption{Ablation experiments for optical flow estimation on the Sintel and KITTI datasets. 
The best configuration is to initialize a ViT-Large encoder with MAE\_st~\cite{feichtenhofer2022masked} and refine for six iterations.
\label{tab:ablations}
}
\end{minipage}
\end{table*}

\subsection{Optical Flow}

\noindent\textbf{Architecture.}
While our framework can operate with any pretrained video transformer, we use MAE\_st ViT-Large~\cite{feichtenhofer2022masked} as the default.
The decoder is implemented as a ConvGRU with six convolutional layers: 
three for horizontal feature extraction and three for vertical feature extraction. 
It takes two inputs: a source image representation and flow features from the previous step (extracted by a motion encoder). 
The flow map is generated by passing the updated hidden state to a flow prediction head (two convolutional layers). 
To achieve full-resolution predictions, 
a mask layer, consisting of two convolutional layers, 
is also trained to indicate how to take the weighted sum of low-resolution neighbors to calculate high-resolution flow.

\noindent\textbf{Metrics.}
We report average endpoint error (EPE) and F1.
The EPE evaluates the average distance in pixels between the predicted and ground truth flow, using the $l_2$ distance. 
Meanwhile, the F1 score measures the percentage of outlier predictions, identified as those with an error exceeding 3 pixels or over 5\% of the magnitude of the ground truth flow.

\subsubsection{Cross Dataset Generalization}
We use the one-cycle learning rate scheduler with cosine annealing.
First, we train the proposed model on Chairs for 40K steps with a batch size of 8 and image size of $368\times 496$,
followed by 400K iterations of training on Things with a batch size of 8 on $384 \times 768$ image crops.
The highest learning rates on these two datasets are $1\times 10^{-4}$ and $2.5\times 10^{-5}$, respectively.
Our augmentation follows Flowformer~\cite{shi2023flowformer++}.
This stage is denoted as ``C+T'',
and then the model is evaluated on the training split of Sintel and KITTI.
A standard tiling scheme~\cite{jaegle2021perceiver, huang2022flowformer, shi2023flowformer++} with a stride step of 224 is used for evaluation.

Table~\ref{tab:optical_flow} presents cross-dataset generalization results with comparisons to prior art.
Our approach surpasses existing alternatives by a large margin on all the datasets.
Compared with the current state-of-the-art SAMFlow which relies on additional large-scale segmentation labels,
GeoViT reduces the error by  20.7\%, 15.6 \% respectively on the training set of Sintel (clean) and Sintel (final).
On KITTI, the EPE and F1 is decreased by 8.4\% and 6.8\%.

\subsubsection{Ablation Study}
Under the same settings used for cross-dataset generalization,
we examine several design choices.

\noindent\textit{Pretrained video foundation models.}
We compare various choices of pretrained video foundation models, including MAE\_st \cite{feichtenhofer2022masked}, spatial MAE \cite{he2022masked}, MVD~\cite{wang2023masked}, InternVideo~\cite{wang2022internvideo} and UMT~\cite{li2023unmasked}.
Table~\ref{tab:abl_pretraining} summarizes the results.
In comparison with other pretrained video models, 
MAE\_st achieves the best performance.
This is potentially because other works which aim to improve semantic capability  
by reconstructing learned features~\cite{wang2023masked}, 
distilling from highly semantic features~\cite{li2023unmasked} such as CLIP~\cite{radford2021learning}, 
or contrasting with language representations~\cite{wang2022internvideo}
do not necessarily translate well for low-level representations, leading to inferior performance.
Improvement over the 2D MAE model pretrained on images suggests that inter-frame pretraining on videos is vital for this two-frame task. 

\noindent\textit{Model Size.}
Table~\ref{tab:abl_model_size} demonstrates the scalability of our adaptation.
The downstream flow results show a consistent improvement when increasing the pre-trained model size.
This observation underscores the potential for further enhancement of our approach with even larger model sizes.

\noindent\textit{Iterative Refinement Steps.}
The results in Table~\ref{tab:abl_iter} show that flow estimation performance generally improves with an increased number of refinement steps.
The performance of the 12-iteration model is slightly better than that of the 6-iteration model, but the difference is minor. Therefore, for efficiency, we choose 6 as the default iteration number.

\noindent\textit{Decoder w.r.t. RAFT.}
Table~\ref{tab:abl_decoder} ablates the role of warping images as opposed to using cost volumes.
We maintain RAFT's cost volume features but use the video foundation model as the context encoder.
Using the video foundation model as the context encoder improves RAFT on Sintel and performs comparably on KITTI.
Our GeoViT outperforms it by a large margin, validating the role of warping images.


\subsubsection{Sintel benchmark}
We evaluate model performance on the Sintel benchmark server.
Following~\cite{xu2023unifying},
we finetune the model on a joint dataset composed of Things, Sintel, KITTI-2015, and HD1K for 200K iterations, with a batch size of 8, image size of $288\times 960$, and highest learning rate of $2.5\times 10^{-5}$.
We proceeded with a Sintel finetuning stage for 5K iterations with a batch size of 8, image size of $416\times 1024$, and highest learning rate of $1\times 10^{-5}$.

As shown in Table~\ref{tab:optical_flow_sintel},
GeoViT sets a new performance standard on the Sintel benchmark.
Our method outperforms the prior state-of-the-art SAM-Flow by 21\% on the clean split and by 9.6\% on the final split, despite being more generic and using less human supervision.

\begin{table}[t]
\footnotesize
\centering
\scriptsize
{
\begin{tabular}[t]{ccccc}
\hline
\multirow{2}{*}{ Training Data } & \multirow{2}{*}{ Method } & \multicolumn{2}{c}{ Sintel (test) } \\
\cmidrule(lr){3-4}
& & Clean & Final \\
\midrule 
\multirow{16}{*}{\shortstack{$\mathrm{C}+\mathrm{T}+\mathrm{S}$\\$+\mathrm{K}+\mathrm{H}$}} & LiteFlowNet2~\cite{hui2020lightweight} & 3.48 & 4.69 \\
& PWC-Net+~\cite{sun2019models} & 3.45 & 4.60 \\
& VCN~\cite{yang2019volumetric} & 2.81 & 4.40 \\
& MaskFlowNet~\cite{zhao2020maskflownet} & 2.52 & 4.17 \\
& S-Flow~\cite{zhang2021separable} & 1.50 & 2.67 \\
& RAFT~\cite{teed2020raft} & 1.94 & 3.18 \\
& RAFT$^*$~\cite{teed2020raft} & 1.61 & 2.86 \\
& FM-RAFT~\cite{jiang2021learning} & 1.72 & 3.60 \\
& GMA~\cite{jiang2021learning} & 1.40 & 2.88 \\
& GMA$^*$~\cite{jiang2021learning} & 1.39 & 2.47 \\
& GMFlow~\cite{xu2022gmflow} & 1.74 & 2.90 \\
& GMFlowNet~\cite{zhao2022global} & 1.39 & 2.65 \\
& CRAFT~\cite{sui2022craft} & 1.45 & 2.42 \\
& SKFlow$^*$~\cite{sun2022skflow} & 1.28 & 2.23 \\
& FlowFormer~\cite{huang2022flowformer} & 1.16 & 2.09 \\
& FlowFormer++~\cite{shi2023flowformer++} & 1.07 & \underline{1.94} \\
& GMFlow+\cite{weinzaepfel2023croco} & 1.03 & 2.12 \\
& SAM-Flow~\cite{zhou2024samflow} & \underline{1.00} & {2.08} \\
& DPFlow~\cite{morimitsu2025dpflow} &1.04&1.97\\
& GeoViT & \textbf{0.79$^{\dagger}$} & \textbf{1.88$^{\dagger}$} \\
\hline
\end{tabular}
}
\caption{ 
Experiments on Sintel~\cite{butler2012naturalistic} benchmark.
`C + K + S + K + H' denotes finetuning on the combined Sintel, KITTI, and HD1K training sets after `C + T' training.
$*$ denotes methods with warm-start proposed in RAFT~\cite{teed2020raft}, 
where the flow is initialized with previous image pair flow estimation.
$^{\dagger}$ represents evaluation with tiling technique~\cite{jaegle2021perceiver}.
Our approach ranks 1st on the Sintel benchmark.
}
\label{tab:optical_flow_sintel}
\end{table}


\subsubsection{KITTI benchmark}
\label{subsec:kitti}
We transfer the Sintel stage-trained model (with the highest learning rate of 2.5e-5) and submit our result to the KITTI evaluation server.
As shown in Table~\ref{tab:optical_flow_kitti}, 
GeoViT achieves an F1-all score of $3.79$. This significantly improves upon SOTAs, reducing the error by 11.2\%.


\subsubsection{Qualitative Comparison}
\begin{figure*}[t]
\begin{center}
   \includegraphics[width=0.9\linewidth]{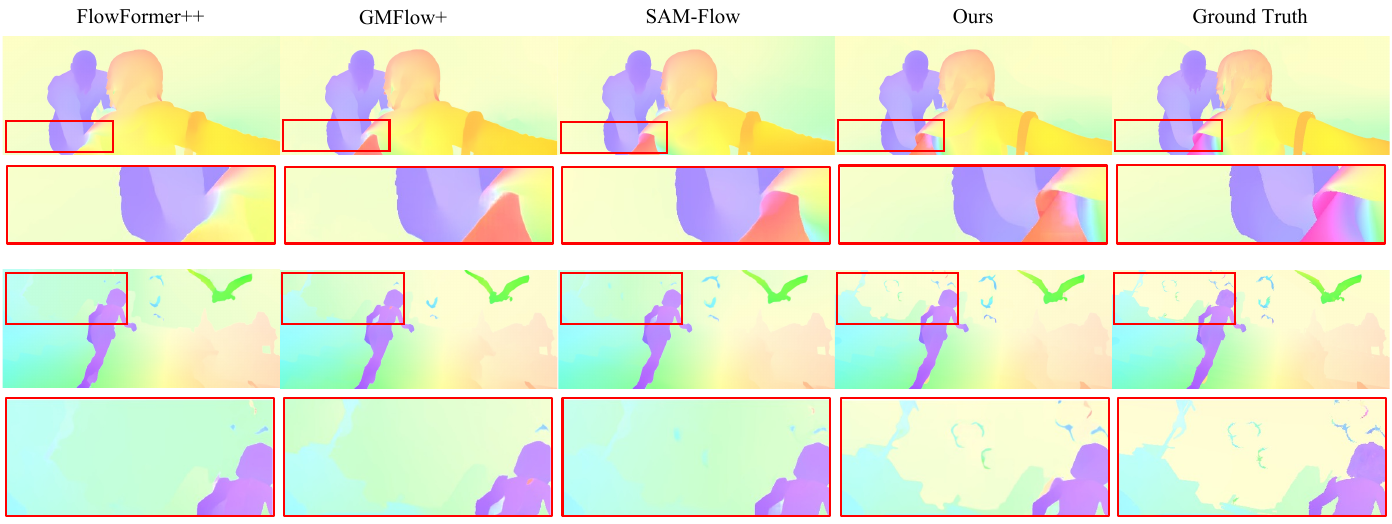}
\end{center}
   \caption{Visualized prediction comparison on Sintel (clean) dataset. 
   Our approach is more accurate with more fine-grained estimates (human shoulder region in case \#1), a higher recall of small objects (bird region in case \#2), and a crisper motion boundary (case \#2). The highlighted region is zoomed in for better visual comparison.
   }
\label{fig:vis_sintel_clean}
\end{figure*}

We compare the visualized flow maps on Sintel (clean) dataset with strong competitors: FlowFormer++, GMFlow+, and SAM-Flow.
As demonstrated in Figure~\ref{fig:vis_sintel_clean},
our predictions show significant improvements.
First, our approach provides more detailed estimates; 
for instance, in case \#1, 
it accurately captures the different motion patterns of the arm and back. 
Second, GeoViT exhibits a higher recall for small objects, 
correctly predicting all bird motions in case \#2.
Lastly, it generates sharper motion boundaries, as seen on the left side of case \#2.



\begin{table}[t]
\footnotesize
\centering
\scriptsize
{
\begin{tabular}[t]{ccc}
\hline
\multirow{2}{*}{ Training Data } & \multirow{2}{*}{ Method }  & KITTI-15 (test)\\
\cmidrule(lr){3-3}& & F1-all \\
\midrule 
\multirow{16}{*}{\shortstack{$\mathrm{C}+\mathrm{T}+\mathrm{S}$\\$+\mathrm{K}+\mathrm{H}$}} & LiteFlowNet2~\cite{hui2020lightweight} & 7.74 \\
& PWC-Net+~\cite{sun2019models} & 7.72 \\
& VCN~\cite{yang2019volumetric} & 6.30 \\
& MaskFlowNet~\cite{zhao2020maskflownet} & 6.10 \\
& S-Flow~\cite{zhang2021separable} & 4.64 \\
& RAFT~\cite{teed2020raft} & 5.10 \\
& RAFT$^*$~\cite{teed2020raft} & 5.10 \\
& FM-RAFT~\cite{jiang2021learning} & 6.17 \\
& GMA~\cite{jiang2021learning} & 5.15 \\
& GMA$^*$~\cite{jiang2021learning} & 5.15 \\
& GMFlow~\cite{xu2022gmflow} & 9.32 \\
& GMFlowNet~\cite{zhao2022global} & 4.79 \\
& CRAFT~\cite{sui2022craft} & 4.79 \\
& SKFlow$^*$~\cite{sun2022skflow} & 4.84 \\
& FlowFormer~\cite{huang2022flowformer} & $4.68^{\dagger}$ \\
& FlowFormer++~\cite{shi2023flowformer++} & $4.52^{\dagger}$ \\
& GMFlow+\cite{weinzaepfel2023croco} & 4.27 \\
& SAM-Flow~\cite{zhou2024samflow} & 4.49 \\
& DPFlow~\cite{morimitsu2025dpflow} & \textbf{3.56}\\
& GeoViT & \underline{3.79}$^{\dagger}$ \\
\hline
\end{tabular}
}
\caption{ 
Experiments on KITTI~\cite{geiger2013vision} benchmark.
`C + K + S + K + H' denotes finetuning on the combined Sintel, KITTI, and HD1K training sets after `C + T' training.
$^{\dagger}$ represents tiling technique~\cite{jaegle2021perceiver}.
}
\label{tab:optical_flow_kitti}
\end{table}

\subsection{Stereo Matching}
\noindent\textbf{Datasets and Evaluation.}
We validate GeoViT on stereo matching with the ETH3D Stereo~\cite{schops2017multi} dataset.
Following~\cite{xu2023unifying}, we conduct two-stage training on the synthetic Scene Flow training set~\cite{mayer2016large}
and a combined dataset: 
Scene Flow~\cite{mayer2016large}, Tartan Air~\cite{wang2020tartanair}, Sintel Stereo~\cite{butler2012naturalistic}, CREStereo Dataset~\cite{li2022practical}, InStereo2K~\cite{bao2020instereo2k} and ETH3D~\cite{schops2017multi}.
We report the performance of GeoViT using the online ETH benchmark~\cite{schops2017multi}.

\noindent\textbf{Training details.}
In the first stage,
GeoViT is initialized with the optical flow parameters trained on ``C+T''.
This approach treats disparity prediction as a one-dimensional optical flow by using the first dimension of the predicted flow values as the output and concatenating zeros to convert the disparity values into two-dimensional values for refinement in the decoder. 
Our model is trained on the Scene Flow dataset for 100k iterations, with a batch size of 8 and an image size of 384$\times$768. 
The maximum learning rate is set to 5e-5.
In the second stage, we finetune our model for an additional 100k iterations, with a batch size of 8 and an image size of 416$\times$640, 
with a maximum learning rate of 5e-6.


\noindent\textbf{Metrics.}
The evaluation uses a widely recognized metric named bad $np$, which measures the percentage of pixels where the absolute error between the predicted disparity and the ground-truth disparity is greater than $np$ pixels. 
We report bad 1.0, bad 2.0, and bad 4.0.


\begin{table}[!t]
    \centering
    \scriptsize
    {
    \begin{tabular}{lccccccccccccc}
    \toprule
    
    Model & bad 1.0 & bad 2.0 & bad 4.0 \\
    
    \midrule
    GANet~\cite{zhang2019ga} & 6.56 & 1.10 & 0.54 \\
    AANet~\cite{xu2020aanet} & 5.01 & 1.66 & 0.75 \\
    CFNet~\cite{shen2021cfnet} & 3.31 & 0.77 & 0.31 \\
    RAFT-Stereo~\cite{lipson2021raft} & 2.44 & 0.44 & 0.15 \\
    CREStereo~\cite{li2022practical} & 0.98 & 0.22 & 0.10 \\
    GMStereo~\cite{xu2023unifying} & 1.83 & 0.25 & 0.08 \\
    MonSter~\cite{cheng2025monster} & \textbf{0.72} & 0.42 & 0.20 \\
    GeoViT & 1.16 & \textbf{0.19} & \textbf{0.03} \\

    \bottomrule
    \end{tabular}
    }
    \caption{Stereo performance on ETH3D stereo test set. 
    Our method achieves the best results on two of the three metrics.}
    \label{tab:stereo_eth3d}
\end{table}

\noindent\textbf{Comparison Results.}
Comparison results are presented in Table~\ref{tab:stereo_eth3d}.
Our model achieves the highest performance in terms of ``bad 2.0'' and ``bad 4.0''.
For ``bad 1.0'', our results are comparable to those of CREStereo~\cite{li2022practical} and significantly outperform the prior unified solution GMStereo~\cite{xu2023unifying} (1.16 vs. 1.83).

\subsection{Depth estimation}
\noindent\textbf{Datasets and Evaluation.}
We perform unrectified stereo depth estimation using a combined dataset consisting of RGBD-SLAM~\cite{sturm2012benchmark}, SUN3D~\cite{xiao2013sun3d}, and Scenes11~\cite{ummenhofer2017demon} for training, and we evaluate on their respective test datasets.
Our model is trained for 100k iterations with a batch size of 8, an image size of 448$\times$576, and the highest learning rate of 2.5e-5.
The iteration number is set to 1 for depth estimation.

\noindent\textbf{Metrics.}
In line with~\cite{tang2018ba, im2019dpsnet, xu2023unifying}, 
we employ four evaluation metrics: Absolute Relative difference (Abs Rel), Squared Relative difference (Sq Rel), Root Mean Squared Error (RMSE), and Root Mean Squared Error in log scale (RMSE log).

\noindent\textbf{Comparison Results.}
As presented in Table~\ref{tab:depth_demon}, our method exhibits performance that is either comparable to or superior to that of prior methods. 
On the RGBD-SLAM, the metrics in which our method underperforms indicate negligible differences, with the largest error difference 
being merely 0.027 in terms of Sq Rel. 
Conversely, our model surpasses the previous best model in terms of RMSE, achieving an improvement of 0.048.
On the SUN3D, our model achieves the best results on 3 of the 4 metrics.
On Scenes11, our model achieves the best performance in 2 of the 4 metrics.
Our visualized failure case analysis reveals that our approach tends to produce overly smooth depth predictions, leading to slightly reduced performance on complex scene datasets like RGBD-SLAM and Scenes11 compared to SUN3D.

\begin{table}[t]
    \centering
    \setlength{\tabcolsep}{1.pt} %
    \scriptsize
    {
    \begin{tabular}{llcccccccccccc}
    \toprule
    
    Dataset & Model & Abs Rel & Sq Rel & RMSE & RMSE log \\
    
    \midrule
    
    \multirow{5}{*}[-2pt]{RGBD-SLAM} & DeMoN~\cite{ummenhofer2017demon} & 0.157 & 0.524 & 1.780 & 0.202  \\
    & DeepMVS~\cite{huang2018deepmvs} & 0.294 & 0.430 & 0.868 & 0.351 \\
    & DPSNet~\cite{im2019dpsnet} & 0.154 & 0.215 & 0.723 & 0.226 &   \\
    & IIB~\cite{yifan2022input} & \textbf{0.095} & - & 0.550 & - &  \\
    & GMDepth~\cite{xu2023unifying} & 0.101 & \textbf{0.177} & 0.556 & \textbf{0.167} \\
    &GeoViT &0.106 & 0.204 & \textbf{0.508} & 0.171\\
    
    \midrule
    
    \multirow{5}{*}[-2pt]{SUN3D} & DeMoN~\cite{ummenhofer2017demon} & 0.214 & 1.120 & 2.421 & 0.206  \\
    & DeepMVS~\cite{huang2018deepmvs} & 0.282 & 0.435 & 0.944 & 0.363 \\
    & DPSNet~\cite{im2019dpsnet} & 0.147 & 0.107 & 0.427 & 0.191 \\
    & IIB~\cite{yifan2022input} & 0.099 & - & \textbf{0.29}3 & - &  \\
    & GMDepth~\cite{xu2023unifying} & 0.112 & \textbf{0.068} & 0.336 & 0.146 \\
    & GeoViT & \textbf{0.095} & \textbf{0.068} & 0.552 & \textbf{0.124} \\
    
    \midrule
    
    \multirow{5}{*}[-2pt]{Scenes11} & DeMoN~\cite{ummenhofer2017demon} & 0.556 & 3.402 & 2.603 & 0.391 \\
    & DeepMVS~\cite{huang2018deepmvs} & 0.210 & 0.373 & 0.891 & 0.270 \\
    & DPSNet~\cite{im2019dpsnet} & 0.056 & 0.144 & 0.714 & 0.140 \\
    & IIB~\cite{yifan2022input} & 0.056 & - & 0.523 & - \\
    & GMDepth~\cite{xu2023unifying} & 0.050 & 0.069 & 0.491 & \textbf{0.106} \\
    & GeoViT & 0.118 & \textbf{0.059} & \textbf{0.318} & 0.146 \\

    \bottomrule
    \end{tabular}
    }
    \caption{Depth performance on DeMoN test datasets. 
    Our approach obtains better or comparable performance.}
    \label{tab:depth_demon}
\end{table}

\section{Conclusion}

In this work, we present a comprehensive study on fine-tuning large-scale pretrained models for multi-view geometry tasks, including optical flow estimation, stereo matching, and unrectified two-view depth estimation.
The core idea is to transfer a general-purpose video foundation model to the domain of geometry estimation with minimal adaptation: by either appending a linear layer as the decoder or embedding it into an iterative refinement loop.
Unlike traditional methods, which often rely on complex, task-specific designs and pretraining strategies, 
our approach is both conceptually straightforward and effective in addressing a wide range of geometric tasks. 
This work not only advances the state of the art in multi-view geometry estimation but also lays the groundwork for future research in leveraging video-pretrained models in this domain.

\bibliography{egbib}

\begin{thebibliography}{79}
\providecommand{\natexlab}[1]{#1}

\bibitem[{Alayrac et~al.(2022)Alayrac, Donahue, Luc, Miech, Barr, Hasson, Lenc, Mensch, Millican, Reynolds et~al.}]{alayrac2022flamingo}
Alayrac, J.-B.; Donahue, J.; Luc, P.; Miech, A.; Barr, I.; Hasson, Y.; Lenc, K.; Mensch, A.; Millican, K.; Reynolds, M.; et~al. 2022.
\newblock Flamingo: a visual language model for few-shot learning.
\newblock \emph{Advances in neural information processing systems}, 35: 23716--23736.

\bibitem[{Bai et~al.(2024)Bai, Geng, Mangalam, Bar, Yuille, Darrell, Malik, and Efros}]{bai2024sequential}
Bai, Y.; Geng, X.; Mangalam, K.; Bar, A.; Yuille, A.~L.; Darrell, T.; Malik, J.; and Efros, A.~A. 2024.
\newblock Sequential modeling enables scalable learning for large vision models.
\newblock In \emph{Proceedings of the IEEE/CVF Conference on Computer Vision and Pattern Recognition}, 22861--22872.

\bibitem[{Bao et~al.(2020)Bao, Wang, Xu, Guo, Hong, and Zhang}]{bao2020instereo2k}
Bao, W.; Wang, W.; Xu, Y.; Guo, Y.; Hong, S.; and Zhang, X. 2020.
\newblock Instereo2k: a large real dataset for stereo matching in indoor scenes.
\newblock \emph{Science China Information Sciences}, 63: 1--11.

\bibitem[{Brown(2020)}]{brown2020language}
Brown, T.~B. 2020.
\newblock Language models are few-shot learners.
\newblock \emph{arXiv preprint arXiv:2005.14165}.

\bibitem[{Butler et~al.(2012)Butler, Wulff, Stanley, and Black}]{butler2012naturalistic}
Butler, D.~J.; Wulff, J.; Stanley, G.~B.; and Black, M.~J. 2012.
\newblock A naturalistic open source movie for optical flow evaluation.
\newblock In \emph{Computer Vision--ECCV 2012: 12th European Conference on Computer Vision, Florence, Italy, October 7-13, 2012, Proceedings, Part VI 12}, 611--625. Springer.

\bibitem[{Chen et~al.(2020)Chen, Kornblith, Norouzi, and Hinton}]{chen2020simple}
Chen, T.; Kornblith, S.; Norouzi, M.; and Hinton, G. 2020.
\newblock A simple framework for contrastive learning of visual representations.
\newblock In \emph{International conference on machine learning}, 1597--1607. PMLR.

\bibitem[{Cheng et~al.(2025)Cheng, Liu, Xu, Wang, Zhang, Deng, Zang, Chen, Cai, and Yang}]{cheng2025monster}
Cheng, J.; Liu, L.; Xu, G.; Wang, X.; Zhang, Z.; Deng, Y.; Zang, J.; Chen, Y.; Cai, Z.; and Yang, X. 2025.
\newblock MonSter: Marry Monodepth to Stereo Unleashes Power.
\newblock In \emph{Proceedings of the IEEE/CVF Conference on Computer Vision and Pattern Recognition}.

\bibitem[{Cheng et~al.(2017)Cheng, Tsai, Wang, and Yang}]{cheng2017segflow}
Cheng, J.; Tsai, Y.-H.; Wang, S.; and Yang, M.-H. 2017.
\newblock Segflow: Joint learning for video object segmentation and optical flow.
\newblock In \emph{Proceedings of the IEEE international conference on computer vision}, 686--695.

\bibitem[{Collins(1996)}]{collins1996space}
Collins, R.~T. 1996.
\newblock A space-sweep approach to true multi-image matching.
\newblock In \emph{Proceedings CVPR IEEE computer society conference on computer vision and pattern recognition}, 358--363. Ieee.

\bibitem[{Ding et~al.(2022)Ding, Yuan, Zhu, Zhang, Liu, Wang, and Liu}]{ding2022transmvsnet}
Ding, Y.; Yuan, W.; Zhu, Q.; Zhang, H.; Liu, X.; Wang, Y.; and Liu, X. 2022.
\newblock Transmvsnet: Global context-aware multi-view stereo network with transformers.
\newblock In \emph{Proceedings of the IEEE/CVF conference on computer vision and pattern recognition}, 8585--8594.

\bibitem[{Dosovitskiy et~al.(2015)Dosovitskiy, Fischer, Ilg, Hausser, Hazirbas, Golkov, Van Der~Smagt, Cremers, and Brox}]{dosovitskiy2015flownet}
Dosovitskiy, A.; Fischer, P.; Ilg, E.; Hausser, P.; Hazirbas, C.; Golkov, V.; Van Der~Smagt, P.; Cremers, D.; and Brox, T. 2015.
\newblock Flownet: Learning optical flow with convolutional networks.
\newblock In \emph{Proceedings of the IEEE international conference on computer vision}, 2758--2766.

\bibitem[{Feichtenhofer et~al.(2022)Feichtenhofer, Li, He et~al.}]{feichtenhofer2022masked}
Feichtenhofer, C.; Li, Y.; He, K.; et~al. 2022.
\newblock Masked autoencoders as spatiotemporal learners.
\newblock \emph{Advances in neural information processing systems}, 35: 35946--35958.

\bibitem[{Geiger et~al.(2013)Geiger, Lenz, Stiller, and Urtasun}]{geiger2013vision}
Geiger, A.; Lenz, P.; Stiller, C.; and Urtasun, R. 2013.
\newblock Vision meets robotics: The kitti dataset.
\newblock \emph{The International Journal of Robotics Research}, 32(11): 1231--1237.

\bibitem[{Gu et~al.(2020)Gu, Fan, Zhu, Dai, Tan, and Tan}]{gu2020cascade}
Gu, X.; Fan, Z.; Zhu, S.; Dai, Z.; Tan, F.; and Tan, P. 2020.
\newblock Cascade cost volume for high-resolution multi-view stereo and stereo matching.
\newblock In \emph{Proceedings of the IEEE/CVF conference on computer vision and pattern recognition}, 2495--2504.

\bibitem[{Guo et~al.(2022)Guo, Li, Yang, Wang, Taylor, Unberath, Yuille, and Li}]{guo2022context}
Guo, W.; Li, Z.; Yang, Y.; Wang, Z.; Taylor, R.~H.; Unberath, M.; Yuille, A.; and Li, Y. 2022.
\newblock Context-enhanced stereo transformer.
\newblock In \emph{European Conference on Computer Vision}, 263--279. Springer.

\bibitem[{He et~al.(2022)He, Chen, Xie, Li, Doll{\'a}r, and Girshick}]{he2022masked}
He, K.; Chen, X.; Xie, S.; Li, Y.; Doll{\'a}r, P.; and Girshick, R. 2022.
\newblock Masked autoencoders are scalable vision learners.
\newblock In \emph{Proceedings of the IEEE/CVF Conference on Computer Vision and Pattern Recognition}, 16000--16009.

\bibitem[{He et~al.(2020)He, Fan, Wu, Xie, and Girshick}]{he2020momentum}
He, K.; Fan, H.; Wu, Y.; Xie, S.; and Girshick, R. 2020.
\newblock Momentum contrast for unsupervised visual representation learning.
\newblock In \emph{Proceedings of the IEEE/CVF conference on computer vision and pattern recognition}, 9729--9738.

\bibitem[{Huang et~al.(2023)Huang, Zhao, Zhang, Qiao, and Wang}]{huang2023mgmae}
Huang, B.; Zhao, Z.; Zhang, G.; Qiao, Y.; and Wang, L. 2023.
\newblock Mgmae: Motion guided masking for video masked autoencoding.
\newblock In \emph{Proceedings of the IEEE/CVF International Conference on Computer Vision}, 13493--13504.

\bibitem[{Huang et~al.(2018)Huang, Matzen, Kopf, Ahuja, and Huang}]{huang2018deepmvs}
Huang, P.-H.; Matzen, K.; Kopf, J.; Ahuja, N.; and Huang, J.-B. 2018.
\newblock Deepmvs: Learning multi-view stereopsis.
\newblock In \emph{Proceedings of the IEEE conference on computer vision and pattern recognition}, 2821--2830.

\bibitem[{Huang et~al.(2022)Huang, Shi, Zhang, Wang, Cheung, Qin, Dai, and Li}]{huang2022flowformer}
Huang, Z.; Shi, X.; Zhang, C.; Wang, Q.; Cheung, K.~C.; Qin, H.; Dai, J.; and Li, H. 2022.
\newblock Flowformer: A transformer architecture for optical flow.
\newblock In \emph{European Conference on Computer Vision}, 668--685. Springer.

\bibitem[{Hui, Tang, and Loy(2018)}]{hui2018liteflownet}
Hui, T.-W.; Tang, X.; and Loy, C.~C. 2018.
\newblock Liteflownet: A lightweight convolutional neural network for optical flow estimation.
\newblock In \emph{Proceedings of the IEEE conference on computer vision and pattern recognition}, 8981--8989.

\bibitem[{Hui, Tang, and Loy(2020)}]{hui2020lightweight}
Hui, T.-W.; Tang, X.; and Loy, C.~C. 2020.
\newblock A lightweight optical flow cnn—revisiting data fidelity and regularization.
\newblock \emph{IEEE transactions on pattern analysis and machine intelligence}, 43(8): 2555--2569.

\bibitem[{Im et~al.(2019)Im, Jeon, Lin, and Kweon}]{im2019dpsnet}
Im, S.; Jeon, H.-G.; Lin, S.; and Kweon, I.~S. 2019.
\newblock Dpsnet: End-to-end deep plane sweep stereo.
\newblock \emph{arXiv preprint arXiv:1905.00538}.

\bibitem[{Jaegle et~al.(2021)Jaegle, Gimeno, Brock, Vinyals, Zisserman, and Carreira}]{jaegle2021perceiver}
Jaegle, A.; Gimeno, F.; Brock, A.; Vinyals, O.; Zisserman, A.; and Carreira, J. 2021.
\newblock Perceiver: General perception with iterative attention.
\newblock In \emph{International conference on machine learning}, 4651--4664. PMLR.

\bibitem[{Jiang et~al.(2021{\natexlab{a}})Jiang, Campbell, Lu, Li, and Hartley}]{jiang2021learning}
Jiang, S.; Campbell, D.; Lu, Y.; Li, H.; and Hartley, R. 2021{\natexlab{a}}.
\newblock Learning to estimate hidden motions with global motion aggregation.
\newblock In \emph{Proceedings of the IEEE/CVF International Conference on Computer Vision}, 9772--9781.

\bibitem[{Jiang et~al.(2021{\natexlab{b}})Jiang, Lu, Li, and Hartley}]{jiang2021learning_2}
Jiang, S.; Lu, Y.; Li, H.; and Hartley, R. 2021{\natexlab{b}}.
\newblock Learning optical flow from a few matches.
\newblock In \emph{Proceedings of the IEEE/CVF conference on computer vision and pattern recognition}, 16592--16600.

\bibitem[{Li et~al.(2022{\natexlab{a}})Li, Wang, Xiong, Cai, Yan, Yang, Liu, Fan, and Liu}]{li2022practical}
Li, J.; Wang, P.; Xiong, P.; Cai, T.; Yan, Z.; Yang, L.; Liu, J.; Fan, H.; and Liu, S. 2022{\natexlab{a}}.
\newblock Practical stereo matching via cascaded recurrent network with adaptive correlation.
\newblock In \emph{Proceedings of the IEEE/CVF conference on computer vision and pattern recognition}, 16263--16272.

\bibitem[{Li et~al.(2023)Li, Wang, Li, Wang, He, Wang, and Qiao}]{li2023unmasked}
Li, K.; Wang, Y.; Li, Y.; Wang, Y.; He, Y.; Wang, L.; and Qiao, Y. 2023.
\newblock Unmasked teacher: Towards training-efficient video foundation models.
\newblock \emph{arXiv preprint arXiv:2303.16058}.

\bibitem[{Li et~al.(2022{\natexlab{b}})Li, Mao, Girshick, and He}]{li2022exploring}
Li, Y.; Mao, H.; Girshick, R.; and He, K. 2022{\natexlab{b}}.
\newblock Exploring plain vision transformer backbones for object detection.
\newblock In \emph{European Conference on Computer Vision}, 280--296. Springer.

\bibitem[{Li et~al.(2021)Li, Liu, Drenkow, Ding, Creighton, Taylor, and Unberath}]{li2021revisiting}
Li, Z.; Liu, X.; Drenkow, N.; Ding, A.; Creighton, F.~X.; Taylor, R.~H.; and Unberath, M. 2021.
\newblock Revisiting stereo depth estimation from a sequence-to-sequence perspective with transformers.
\newblock In \emph{Proceedings of the IEEE/CVF international conference on computer vision}, 6197--6206.

\bibitem[{Lin et~al.(2023)Lin, Yuan, Zhang, Li, Zheng, and Hu}]{lin2023detr}
Lin, Y.; Yuan, Y.; Zhang, Z.; Li, C.; Zheng, N.; and Hu, H. 2023.
\newblock Detr does not need multi-scale or locality design.
\newblock In \emph{Proceedings of the IEEE/CVF International Conference on Computer Vision}, 6545--6554.

\bibitem[{Lipson, Teed, and Deng(2021)}]{lipson2021raft}
Lipson, L.; Teed, Z.; and Deng, J. 2021.
\newblock Raft-stereo: Multilevel recurrent field transforms for stereo matching.
\newblock In \emph{2021 International Conference on 3D Vision (3DV)}, 218--227. IEEE.

\bibitem[{Liu et~al.(2024)Liu, Li, Wu, and Lee}]{liu2024visual}
Liu, H.; Li, C.; Wu, Q.; and Lee, Y.~J. 2024.
\newblock Visual instruction tuning.
\newblock \emph{Advances in neural information processing systems}, 36.

\bibitem[{Liu et~al.(2023)Liu, Xu, Bertasius, and Niethammer}]{liu2023simpleclick}
Liu, Q.; Xu, Z.; Bertasius, G.; and Niethammer, M. 2023.
\newblock Simpleclick: Interactive image segmentation with simple vision transformers.
\newblock In \emph{Proceedings of the IEEE/CVF International Conference on Computer Vision}, 22290--22300.

\bibitem[{Mayer et~al.(2016)Mayer, Ilg, Hausser, Fischer, Cremers, Dosovitskiy, and Brox}]{mayer2016large}
Mayer, N.; Ilg, E.; Hausser, P.; Fischer, P.; Cremers, D.; Dosovitskiy, A.; and Brox, T. 2016.
\newblock A large dataset to train convolutional networks for disparity, optical flow, and scene flow estimation.
\newblock In \emph{Proceedings of the IEEE conference on computer vision and pattern recognition}, 4040--4048.

\bibitem[{Morimitsu et~al.(2025)Morimitsu, Zhu, Cesar~Jr, Ji, and Yin}]{morimitsu2025dpflow}
Morimitsu, H.; Zhu, X.; Cesar~Jr, R.~M.; Ji, X.; and Yin, X.-C. 2025.
\newblock DPFlow: Adaptive Optical Flow Estimation with a Dual-Pyramid Framework.
\newblock In \emph{arXiv preprint arXiv:2503.14880}.

\bibitem[{Radford et~al.(2021)Radford, Kim, Hallacy, Ramesh, Goh, Agarwal, Sastry, Askell, Mishkin, Clark et~al.}]{radford2021learning}
Radford, A.; Kim, J.~W.; Hallacy, C.; Ramesh, A.; Goh, G.; Agarwal, S.; Sastry, G.; Askell, A.; Mishkin, P.; Clark, J.; et~al. 2021.
\newblock Learning transferable visual models from natural language supervision.
\newblock In \emph{International conference on machine learning}, 8748--8763. PMLR.

\bibitem[{Ranftl, Bochkovskiy, and Koltun(2021)}]{ranftl2021vision}
Ranftl, R.; Bochkovskiy, A.; and Koltun, V. 2021.
\newblock Vision transformers for dense prediction.
\newblock In \emph{Proceedings of the IEEE/CVF international conference on computer vision}, 12179--12188.

\bibitem[{Schops et~al.(2017)Schops, Schonberger, Galliani, Sattler, Schindler, Pollefeys, and Geiger}]{schops2017multi}
Schops, T.; Schonberger, J.~L.; Galliani, S.; Sattler, T.; Schindler, K.; Pollefeys, M.; and Geiger, A. 2017.
\newblock A multi-view stereo benchmark with high-resolution images and multi-camera videos.
\newblock In \emph{Proceedings of the IEEE conference on computer vision and pattern recognition}, 3260--3269.

\bibitem[{Schulter et~al.(2017)Schulter, Vernaza, Choi, and Chandraker}]{schulter2017deep}
Schulter, S.; Vernaza, P.; Choi, W.; and Chandraker, M. 2017.
\newblock Deep network flow for multi-object tracking.
\newblock In \emph{Proceedings of the IEEE Conference on Computer Vision and Pattern Recognition}, 6951--6960.

\bibitem[{Shen, Dai, and Rao(2021)}]{shen2021cfnet}
Shen, Z.; Dai, Y.; and Rao, Z. 2021.
\newblock Cfnet: Cascade and fused cost volume for robust stereo matching.
\newblock In \emph{Proceedings of the IEEE/CVF conference on computer vision and pattern recognition}, 13906--13915.

\bibitem[{Shi et~al.(2023)Shi, Huang, Li, Zhang, Cheung, See, Qin, Dai, and Li}]{shi2023flowformer++}
Shi, X.; Huang, Z.; Li, D.; Zhang, M.; Cheung, K.~C.; See, S.; Qin, H.; Dai, J.; and Li, H. 2023.
\newblock Flowformer++: Masked cost volume autoencoding for pretraining optical flow estimation.
\newblock In \emph{Proceedings of the IEEE/CVF Conference on Computer Vision and Pattern Recognition}, 1599--1610.

\bibitem[{Simonyan and Zisserman(2014)}]{simonyan2014two}
Simonyan, K.; and Zisserman, A. 2014.
\newblock Two-stream convolutional networks for action recognition in videos.
\newblock \emph{Advances in neural information processing systems}, 27.

\bibitem[{Sturm et~al.(2012)Sturm, Engelhard, Endres, Burgard, and Cremers}]{sturm2012benchmark}
Sturm, J.; Engelhard, N.; Endres, F.; Burgard, W.; and Cremers, D. 2012.
\newblock A benchmark for the evaluation of RGB-D SLAM systems.
\newblock In \emph{2012 IEEE/RSJ international conference on intelligent robots and systems}, 573--580. IEEE.

\bibitem[{Sui et~al.(2022)Sui, Li, Geng, Wu, Xu, Liu, Goh, and Zhu}]{sui2022craft}
Sui, X.; Li, S.; Geng, X.; Wu, Y.; Xu, X.; Liu, Y.; Goh, R.; and Zhu, H. 2022.
\newblock Craft: Cross-attentional flow transformer for robust optical flow.
\newblock In \emph{Proceedings of the IEEE/CVF conference on Computer Vision and Pattern Recognition}, 17602--17611.

\bibitem[{Sun et~al.(2018{\natexlab{a}})Sun, Yang, Liu, and Kautz}]{sun2018pwc}
Sun, D.; Yang, X.; Liu, M.-Y.; and Kautz, J. 2018{\natexlab{a}}.
\newblock Pwc-net: Cnns for optical flow using pyramid, warping, and cost volume.
\newblock In \emph{Proceedings of the IEEE conference on computer vision and pattern recognition}, 8934--8943.

\bibitem[{Sun et~al.(2019)Sun, Yang, Liu, and Kautz}]{sun2019models}
Sun, D.; Yang, X.; Liu, M.-Y.; and Kautz, J. 2019.
\newblock Models matter, so does training: An empirical study of cnns for optical flow estimation.
\newblock \emph{IEEE transactions on pattern analysis and machine intelligence}, 42(6): 1408--1423.

\bibitem[{Sun et~al.(2022)Sun, Chen, Zhu, Guo, and Li}]{sun2022skflow}
Sun, S.; Chen, Y.; Zhu, Y.; Guo, G.; and Li, G. 2022.
\newblock Skflow: Learning optical flow with super kernels.
\newblock \emph{Advances in Neural Information Processing Systems}, 35: 11313--11326.

\bibitem[{Sun et~al.(2018{\natexlab{b}})Sun, Kuang, Sheng, Ouyang, and Zhang}]{sun2018optical}
Sun, S.; Kuang, Z.; Sheng, L.; Ouyang, W.; and Zhang, W. 2018{\natexlab{b}}.
\newblock Optical flow guided feature: A fast and robust motion representation for video action recognition.
\newblock In \emph{Proceedings of the IEEE conference on computer vision and pattern recognition}, 1390--1399.

\bibitem[{Tang and Tan(2018)}]{tang2018ba}
Tang, C.; and Tan, P. 2018.
\newblock BA-Net: Dense Bundle Adjustment Networks.
\newblock In \emph{International Conference on Learning Representations}.

\bibitem[{Teed and Deng(2020)}]{teed2020raft}
Teed, Z.; and Deng, J. 2020.
\newblock Raft: Recurrent all-pairs field transforms for optical flow.
\newblock In \emph{Computer Vision--ECCV 2020: 16th European Conference, Glasgow, UK, August 23--28, 2020, Proceedings, Part II 16}, 402--419. Springer.

\bibitem[{Tong et~al.(2022)Tong, Song, Wang, and Wang}]{tong2022videomae}
Tong, Z.; Song, Y.; Wang, J.; and Wang, L. 2022.
\newblock Videomae: Masked autoencoders are data-efficient learners for self-supervised video pre-training.
\newblock \emph{arXiv preprint arXiv:2203.12602}.

\bibitem[{Ummenhofer et~al.(2017)Ummenhofer, Zhou, Uhrig, Mayer, Ilg, Dosovitskiy, and Brox}]{ummenhofer2017demon}
Ummenhofer, B.; Zhou, H.; Uhrig, J.; Mayer, N.; Ilg, E.; Dosovitskiy, A.; and Brox, T. 2017.
\newblock Demon: Depth and motion network for learning monocular stereo.
\newblock In \emph{Proceedings of the IEEE conference on computer vision and pattern recognition}, 5038--5047.

\bibitem[{Wang et~al.(2023{\natexlab{a}})Wang, Huang, Zhao, Tong, He, Wang, Wang, and Qiao}]{wang2023videomae}
Wang, L.; Huang, B.; Zhao, Z.; Tong, Z.; He, Y.; Wang, Y.; Wang, Y.; and Qiao, Y. 2023{\natexlab{a}}.
\newblock Videomae v2: Scaling video masked autoencoders with dual masking.
\newblock In \emph{Proceedings of the IEEE/CVF Conference on Computer Vision and Pattern Recognition}, 14549--14560.

\bibitem[{Wang et~al.(2023{\natexlab{b}})Wang, Chen, Wu, Chen, Dai, Liu, Yuan, and Jiang}]{wang2023masked}
Wang, R.; Chen, D.; Wu, Z.; Chen, Y.; Dai, X.; Liu, M.; Yuan, L.; and Jiang, Y.-G. 2023{\natexlab{b}}.
\newblock Masked video distillation: Rethinking masked feature modeling for self-supervised video representation learning.
\newblock In \emph{Proceedings of the IEEE/CVF Conference on Computer Vision and Pattern Recognition}, 6312--6322.

\bibitem[{Wang et~al.(2023{\natexlab{c}})Wang, Dai, Chen, Huang, Li, Zhu, Hu, Lu, Lu, Li et~al.}]{wang2023internimage}
Wang, W.; Dai, J.; Chen, Z.; Huang, Z.; Li, Z.; Zhu, X.; Hu, X.; Lu, T.; Lu, L.; Li, H.; et~al. 2023{\natexlab{c}}.
\newblock Internimage: Exploring large-scale vision foundation models with deformable convolutions.
\newblock In \emph{Proceedings of the IEEE/CVF conference on computer vision and pattern recognition}, 14408--14419.

\bibitem[{Wang et~al.(2020)Wang, Zhu, Wang, Hu, Qiu, Wang, Hu, Kapoor, and Scherer}]{wang2020tartanair}
Wang, W.; Zhu, D.; Wang, X.; Hu, Y.; Qiu, Y.; Wang, C.; Hu, Y.; Kapoor, A.; and Scherer, S. 2020.
\newblock Tartanair: A dataset to push the limits of visual slam.
\newblock In \emph{2020 IEEE/RSJ International Conference on Intelligent Robots and Systems (IROS)}, 4909--4916. IEEE.

\bibitem[{Wang et~al.(2022)Wang, Li, Li, He, Huang, Zhao, Zhang, Xu, Liu, Wang et~al.}]{wang2022internvideo}
Wang, Y.; Li, K.; Li, Y.; He, Y.; Huang, B.; Zhao, Z.; Zhang, H.; Xu, J.; Liu, Y.; Wang, Z.; et~al. 2022.
\newblock Internvideo: General video foundation models via generative and discriminative learning.
\newblock \emph{arXiv preprint arXiv:2212.03191}.

\bibitem[{Weinzaepfel et~al.(2022)Weinzaepfel, Leroy, Lucas, Br{\'e}gier, Cabon, Arora, Antsfeld, Chidlovskii, Csurka, and Revaud}]{weinzaepfel2022croco}
Weinzaepfel, P.; Leroy, V.; Lucas, T.; Br{\'e}gier, R.; Cabon, Y.; Arora, V.; Antsfeld, L.; Chidlovskii, B.; Csurka, G.; and Revaud, J. 2022.
\newblock CroCo: Self-Supervised Pre-training for 3D Vision Tasks by Cross-View Completion.
\newblock \emph{Advances in Neural Information Processing Systems}, 35: 3502--3516.

\bibitem[{Weinzaepfel et~al.(2023)Weinzaepfel, Lucas, Leroy, Cabon, Arora, Br{\'e}gier, Csurka, Antsfeld, Chidlovskii, and Revaud}]{weinzaepfel2023croco}
Weinzaepfel, P.; Lucas, T.; Leroy, V.; Cabon, Y.; Arora, V.; Br{\'e}gier, R.; Csurka, G.; Antsfeld, L.; Chidlovskii, B.; and Revaud, J. 2023.
\newblock CroCo v2: Improved Cross-view Completion Pre-training for Stereo Matching and Optical Flow.
\newblock In \emph{Proceedings of the IEEE/CVF International Conference on Computer Vision}, 17969--17980.

\bibitem[{Wu et~al.(2018)Wu, Xiong, Yu, and Lin}]{wu2018unsupervised}
Wu, Z.; Xiong, Y.; Yu, S.~X.; and Lin, D. 2018.
\newblock Unsupervised feature learning via non-parametric instance discrimination.
\newblock In \emph{Proceedings of the IEEE conference on computer vision and pattern recognition}, 3733--3742.

\bibitem[{Xiao et~al.(2018)Xiao, Feng, Lin, Liu, and Zhang}]{xiao2018monet}
Xiao, H.; Feng, J.; Lin, G.; Liu, Y.; and Zhang, M. 2018.
\newblock Monet: Deep motion exploitation for video object segmentation.
\newblock In \emph{Proceedings of the IEEE Conference on Computer Vision and Pattern Recognition}, 1140--1148.

\bibitem[{Xiao, Owens, and Torralba(2013)}]{xiao2013sun3d}
Xiao, J.; Owens, A.; and Torralba, A. 2013.
\newblock Sun3d: A database of big spaces reconstructed using sfm and object labels.
\newblock In \emph{Proceedings of the IEEE international conference on computer vision}, 1625--1632.

\bibitem[{Xu et~al.(2023{\natexlab{a}})Xu, Wang, Ding, and Yang}]{xu2023iterative}
Xu, G.; Wang, X.; Ding, X.; and Yang, X. 2023{\natexlab{a}}.
\newblock Iterative geometry encoding volume for stereo matching.
\newblock In \emph{Proceedings of the IEEE/CVF Conference on Computer Vision and Pattern Recognition}, 21919--21928.

\bibitem[{Xu and Zhang(2020)}]{xu2020aanet}
Xu, H.; and Zhang, J. 2020.
\newblock Aanet: Adaptive aggregation network for efficient stereo matching.
\newblock In \emph{Proceedings of the IEEE/CVF conference on computer vision and pattern recognition}, 1959--1968.

\bibitem[{Xu et~al.(2022{\natexlab{a}})Xu, Zhang, Cai, Rezatofighi, and Tao}]{xu2022gmflow}
Xu, H.; Zhang, J.; Cai, J.; Rezatofighi, H.; and Tao, D. 2022{\natexlab{a}}.
\newblock Gmflow: Learning optical flow via global matching.
\newblock In \emph{Proceedings of the IEEE/CVF conference on computer vision and pattern recognition}, 8121--8130.

\bibitem[{Xu et~al.(2023{\natexlab{b}})Xu, Zhang, Cai, Rezatofighi, Yu, Tao, and Geiger}]{xu2023unifying}
Xu, H.; Zhang, J.; Cai, J.; Rezatofighi, H.; Yu, F.; Tao, D.; and Geiger, A. 2023{\natexlab{b}}.
\newblock Unifying flow, stereo and depth estimation.
\newblock \emph{IEEE Transactions on Pattern Analysis and Machine Intelligence}.

\bibitem[{Xu et~al.(2022{\natexlab{b}})Xu, Zhang, Zhang, and Tao}]{xu2022vitpose}
Xu, Y.; Zhang, J.; Zhang, Q.; and Tao, D. 2022{\natexlab{b}}.
\newblock Vitpose: Simple vision transformer baselines for human pose estimation.
\newblock \emph{Advances in Neural Information Processing Systems}, 35: 38571--38584.

\bibitem[{Xu et~al.(2018)Xu, Fu, Yang, and Lee}]{xu2018dynamic}
Xu, Y.-S.; Fu, T.-J.; Yang, H.-K.; and Lee, C.-Y. 2018.
\newblock Dynamic video segmentation network.
\newblock In \emph{Proceedings of the IEEE conference on computer vision and pattern recognition}, 6556--6565.

\bibitem[{Yang and Ramanan(2019)}]{yang2019volumetric}
Yang, G.; and Ramanan, D. 2019.
\newblock Volumetric correspondence networks for optical flow.
\newblock \emph{Advances in neural information processing systems}, 32.

\bibitem[{Yang et~al.(2020)Yang, Mao, Alvarez, and Liu}]{yang2020cost}
Yang, J.; Mao, W.; Alvarez, J.~M.; and Liu, M. 2020.
\newblock Cost volume pyramid based depth inference for multi-view stereo.
\newblock In \emph{Proceedings of the IEEE/CVF conference on computer vision and pattern recognition}, 4877--4886.

\bibitem[{Yifan et~al.(2022)Yifan, Doersch, Arandjelovi{\'c}, Carreira, and Zisserman}]{yifan2022input}
Yifan, W.; Doersch, C.; Arandjelovi{\'c}, R.; Carreira, J.; and Zisserman, A. 2022.
\newblock Input-level inductive biases for 3d reconstruction.
\newblock In \emph{Proceedings of the IEEE/CVF Conference on Computer Vision and Pattern Recognition}, 6176--6186.

\bibitem[{Yin, Darrell, and Yu(2019)}]{yin2019hierarchical}
Yin, Z.; Darrell, T.; and Yu, F. 2019.
\newblock Hierarchical discrete distribution decomposition for match density estimation.
\newblock In \emph{Proceedings of the IEEE/CVF conference on computer vision and pattern recognition}, 6044--6053.

\bibitem[{Zhang et~al.(2019)Zhang, Prisacariu, Yang, and Torr}]{zhang2019ga}
Zhang, F.; Prisacariu, V.; Yang, R.; and Torr, P.~H. 2019.
\newblock Ga-net: Guided aggregation net for end-to-end stereo matching.
\newblock In \emph{Proceedings of the IEEE/CVF conference on computer vision and pattern recognition}, 185--194.

\bibitem[{Zhang et~al.(2021)Zhang, Woodford, Prisacariu, and Torr}]{zhang2021separable}
Zhang, F.; Woodford, O.~J.; Prisacariu, V.~A.; and Torr, P.~H. 2021.
\newblock Separable flow: Learning motion cost volumes for optical flow estimation.
\newblock In \emph{Proceedings of the IEEE/CVF international conference on computer vision}, 10807--10817.

\bibitem[{Zhao et~al.(2020)Zhao, Sheng, Dong, Chang, Xu et~al.}]{zhao2020maskflownet}
Zhao, S.; Sheng, Y.; Dong, Y.; Chang, E.~I.; Xu, Y.; et~al. 2020.
\newblock Maskflownet: Asymmetric feature matching with learnable occlusion mask.
\newblock In \emph{Proceedings of the IEEE/CVF conference on computer vision and pattern recognition}, 6278--6287.

\bibitem[{Zhao et~al.(2022)Zhao, Zhao, Zhang, Zhou, and Metaxas}]{zhao2022global}
Zhao, S.; Zhao, L.; Zhang, Z.; Zhou, E.; and Metaxas, D. 2022.
\newblock Global matching with overlapping attention for optical flow estimation.
\newblock In \emph{Proceedings of the IEEE/CVF Conference on Computer Vision and Pattern Recognition}, 17592--17601.

\bibitem[{Zheng et~al.(2021)Zheng, Ma, Peng, and Yang}]{zheng2021learning}
Zheng, J.; Ma, C.; Peng, H.; and Yang, X. 2021.
\newblock Learning to track objects from unlabeled videos.
\newblock In \emph{Proceedings of the IEEE/CVF international conference on computer vision}, 13546--13555.

\bibitem[{Zhou et~al.(2024)Zhou, He, Tan, and Yan}]{zhou2024samflow}
Zhou, S.; He, R.; Tan, W.; and Yan, B. 2024.
\newblock Samflow: Eliminating any fragmentation in optical flow with segment anything model.
\newblock In \emph{Proceedings of the AAAI Conference on Artificial Intelligence}, volume~38, 7695--7703.

\end{thebibliography}

\clearpage
\includepdf[pages=-]{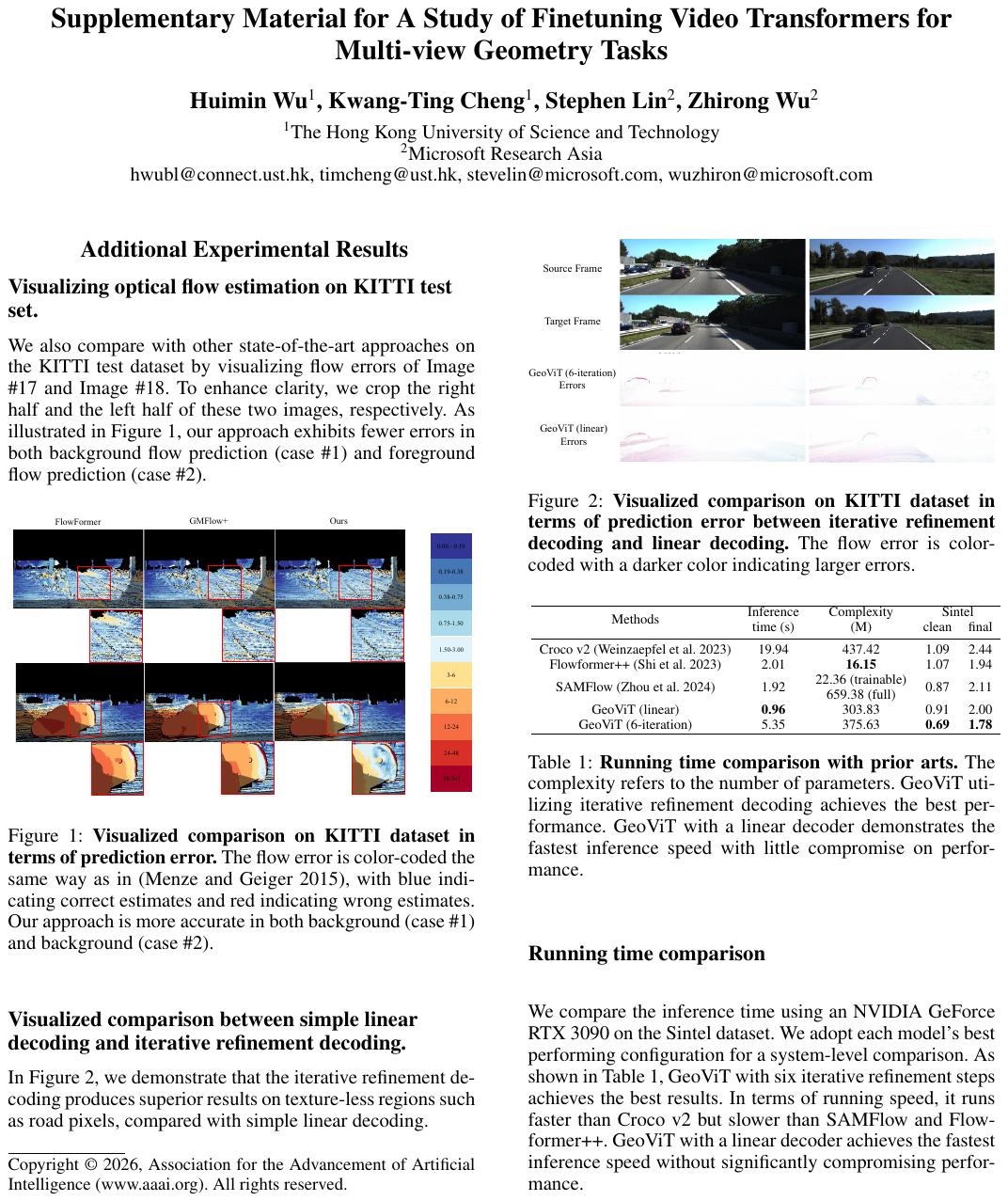}


\end{document}